\documentclass[10pt,twocolumn,letterpaper]{article}

\usepackage{cvpr}              %
\usepackage[accsupp]{axessibility}  %

\usepackage{dblfloatfix}

\definecolor{cvprblue}{rgb}{0.21,0.49,0.74}
\usepackage[pagebackref,breaklinks,colorlinks,allcolors=cvprblue]{hyperref}

\def\paperID{} %
\def\confName{CVPR}
\def\confYear{2026}

\title{Parallel Jacobi Decoding for Fast Autoregressive Image Generation}

\author{
Boya Liao, Ying Li, Siyong Jian, Huan Wang$^{*}$\\
Westlake University\\
\url{https://boyaliao.github.io/PJD/}
}
\usepackage{multirow}
\usepackage{algorithm}
\usepackage{algpseudocode}
\usepackage{amsmath}
\usepackage{amssymb}
\usepackage{xcolor}

\usepackage{float}

\algrenewcommand\alglinenumber[1]{#1:}
\algrenewcommand\algorithmicwhile{\textbf{while}}
\algrenewcommand\algorithmicfor{\textbf{for}}
\algrenewcommand\algorithmicif{\textbf{if}}
\algrenewcommand\algorithmicelse{\textbf{else}}
\algrenewcommand\algorithmicdo{\textbf{do}}
\algrenewcommand\algorithmicend{\textbf{end}}
\algrenewcommand\algorithmicthen{:}

\begin{document}
\renewcommand{\thefootnote}{\fnsymbol{footnote}}
\twocolumn[{%
    \renewcommand\twocolumn[1][]{#1}%
    \maketitle
    \begin{center}
        \vspace{-1.5em}
        \centering
            \includegraphics[width=0.9\linewidth]{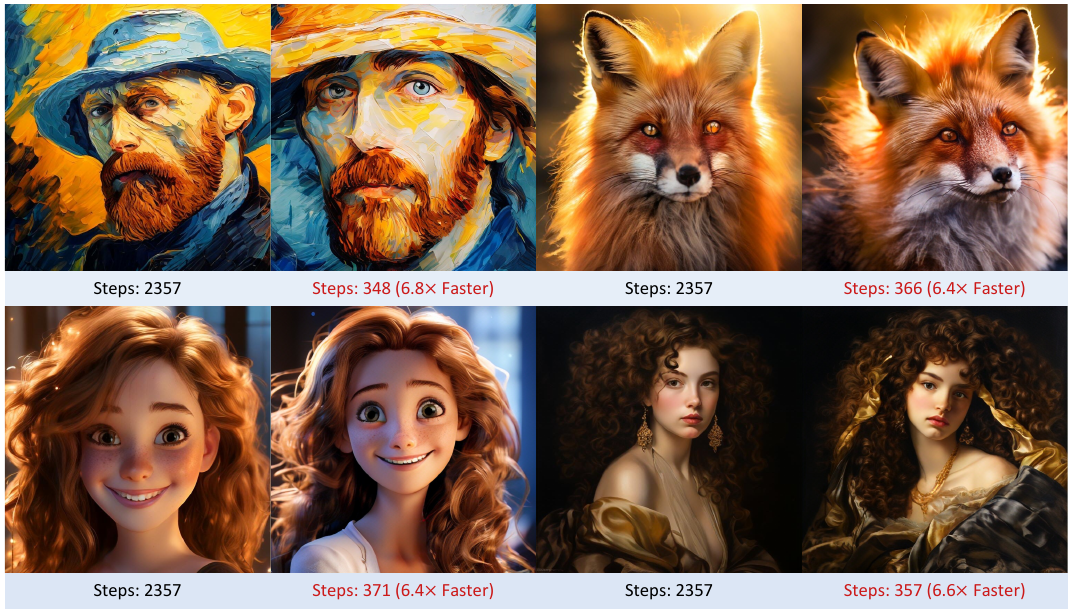}
        
            \captionof{figure}{Images generated by Lumina-mGPT ~\citep{liu2024lumina} using vanilla autoregressive decoding (left) and our Parallel Jacobi Decoding (right). Our method reduces the required autoregressive steps by up to \textbf{6.8×} while maintaining visual fidelity. Prompts for all examples are provided in the supplementary material.}
            \label{fig:teaser}
    \end{center}%
}]
\footnotetext[1]{Corresponding author: \texttt{wanghuan@westlake.edu.cn}}

\begin{abstract}
Autoregressive (AR) models have demonstrated remarkable performance in generating high-fidelity images.
However, their inherently sequential next-token prediction leads to significantly slower inference.
Recent studies have introduced Jacobi-style decoding to accelerate autoregressive image generation.
Extending the draft sequence initially improves efficiency, yet the acceleration quickly saturates as error propagation in the one-dimensional sequence hinders convergence.
Observing that images exhibit strong local spatial correlations, we propose Parallel Jacobi Decoding (PJD), a training-free decoding approach that expands draft tokens in the two-dimensional spatial domain to enable efficient spatially parallel refinement.
PJD adjusts the attention mask to mitigate error accumulation and improve convergence stability.
Extensive experiments on diverse datasets show that PJD achieves \textbf{4.8×–6.4×} acceleration across multiple autoregressive image generation models while maintaining competitive generation quality.
\end{abstract}
    
\section{Introduction}
\label{sec:intro}
Recent advances in large language models (LLMs)~~\citep{achiam2023gpt, team2023gemini, liu2024deepseek, grattafiori2024llama,yang2025qwen3} 
have highlighted the remarkable effectiveness of the autoregressive (AR) paradigm in sequence modeling.
Building on this success, recent studies~~\citep{team2024chameleon, liu2024lumina, sun2024autoregressive, chen2025janus, chern2024anole} have extended AR modeling to image generation, achieving image quality comparable to diffusion-based approaches~~\citep{ho2020denoising, song2020denoising, song2020score, rombach2022high,peebles2023scalable}. 
The transformer architecture underlying AR models allows more flexible composition and offers a unified framework that connects language and vision domains.

Despite these advantages, AR models face a bottleneck arising from their sequential generation process ~\citep{vaswani2017attention, stern2018blockwise, santilli2023accelerating, leviathan2023fast}. Since each forward pass produces only one token, generating an image may require hundreds or even thousands of iterations. This makes inference prohibitively slow for real-time applications.
Although many acceleration techniques ~\citep{salimans2022progressive, lu2022dpm, song2023consistency, luo2023latent,sauer2024adversarial} have been developed for diffusion models, they cannot be applied to AR models because of fundamental differences in their generation mechanisms.

To accelerate AR image generation, recent work has explored adapting acceleration strategies from LLMs to vision.
Speculative decoding~\citep{leviathan2023fast, chen2023accelerating, li2024eagle, cai2024medusa, jang2025lantern} employs a lightweight draft model to propose tokens that are then verified by the main model. Nonetheless, it requires training an auxiliary draft model, which introduces additional computational overhead.
Jacobi decoding~\citep{santilli2023accelerating, kou2024cllms, tengaccelerating, so2025grouped} removes the auxiliary model and iteratively refines a set of candidate tokens until convergence. 
Extending the candidate sequence accelerates inference at first, as multiple tokens can be accepted per iteration. However, the acceleration soon saturates, as later tokens inevitably attend to an increasing number of uncertain candidate tokens, ultimately making refinement harder and convergence significantly slower.

\begin{figure}
  \centering
   \includegraphics[width=1.0\linewidth]{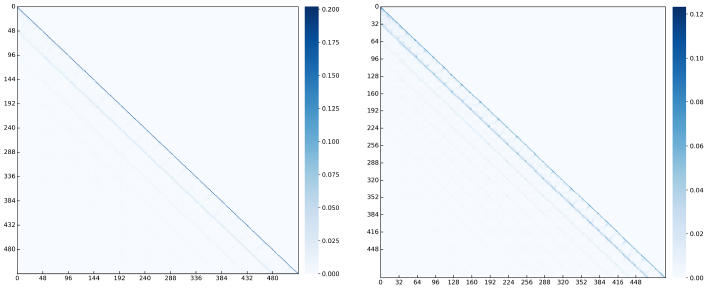}

   \caption{Visualization of the attention maps of Lumina-mGPT~\citep{liu2024lumina} (left) and LlamaGen~\citep{sun2024autoregressive} (right), averaged over all layers and shown for the first 500 tokens. The diagonal and nearby bands indicate that attention mainly concentrates on the current row of each token and neighboring regions in preceding rows, revealing strong local attention patterns.}
   \label{fig:attention-pattern}
\end{figure}

This saturation indicates that the limited scalability of Jacobi decoding originates from its inherently one-dimensional (1D) expansion scheme. 
Unlike text sequences, where each token captures long-range dependencies across the entire context, AR image generation exhibits strong spatial locality, as shown in Figure~\ref{fig:attention-pattern}, where each token mainly attends to its nearby region.
\textit{This observation suggests that the one-dimensional expansion scheme used in conventional Jacobi decoding is suboptimal for images}. Aligning decoding with the two-dimensional (2D) spatial structure may therefore enable more efficient generation.

Motivated by this insight, we propose \textbf{Parallel Jacobi Decoding (PJD)}, a two-dimensional extension of Jacobi decoding for efficient AR image generation.
After the current row provides sufficient context, PJD initializes draft tokens in the next row, enabling concurrent refinement across spatial regions within each iteration.
Because each token primarily attends to generated neighboring tokens, this localized refinement preserves the autoregressive dependency structure and reduces long-range error propagation.
This design allows more tokens to be accepted per iteration, achieving faster and more stable convergence without requiring any additional model training.

We evaluate PJD on two representative AR image models, Lumina-mGPT~\citep{liu2024lumina} and LlamaGen~\citep{sun2024autoregressive}.
Across diverse datasets~\citep{yu2022scaling,lin2014microsoft} and resolutions, PJD consistently accelerates inference while maintaining competitive generation quality.
It achieves \textbf{6.4×} speedup on Lumina-mGPT and \textbf{4.8×} on LlamaGen.
The speedup grows with image resolution, confirming the benefit of two-dimensional spatial parallelism for high-resolution image generation.

In summary, our main contributions are:
\begin{itemize}
  \item We identify that the one-dimensional Jacobi decoding scheme underutilizes the spatial structure of images, limiting scalability in autoregressive image generation.

  \item We introduce a training-free two-dimensional parallel Jacobi decoding framework. Once sufficient spatial context is available, it initializes draft tokens for the next row. By enabling the refinement of multiple rows in one iteration, the framework achieves substantial acceleration.

  \item Experiments on Lumina-mGPT and LlamaGen show that PJD achieves up to \textbf{6.4×} and \textbf{4.8×} speedup, respectively, without compromising image quality.
  Moreover, PJD outperforms other state-of-the-art methods.
\end{itemize}

\section{Related Work}
\label{sec:formatting}
\subsection{Autoregressive Image Generation}
Early AR image generation models such as PixelRNN~\citep{van2016pixel, chen2018pixelsnail} and PixelCNN~\citep{ van2016conditional, kolesnikov2017pixelcnn, salimans2017pixelcnn++} operate directly in pixel space, modeling images via a strictly causal factorization over pixel intensities. These models offer tractable likelihoods and stable training, yet their strictly sequential, pixel-level factorization leads to long dependency chains.

To address this limitation, subsequent work shifted toward token-based AR formulations, in which images are encoded into discrete latent tokens via vector quantization. Early approaches such as VQ-VAE~\citep{oord2017vqvae, razavi2019generating} and VQGAN~\citep{esser2021taming} compressed images into semantically meaningful discrete latent spaces, substantially reducing sequence length and making autoregressive modeling more practical. Building on this paradigm, visual AR transformers have demonstrated strong image generation performance. DALL·E~\citep{ramesh2021zero,ramesh2022hierarchical} showed the feasibility of generating images from text with autoregressive transformers over discrete visual tokens, while Parti~\citep{yu2022scaling} and LlamaGen~\citep{sun2024autoregressive} further improved generation quality through model scaling, improved visual tokenization, and higher-quality training data. Recent works such as Chameleon~\citep{team2024chameleon}, Anole~\citep{chern2024anole}, Lumina-mGPT~\citep{liu2024lumina, xin2025lumina} and Janus-Pro~\citep{chen2025janus} extend token-based autoregressive modeling from visual generation to unified multimodal understanding and generation. Despite these advances in representation and modeling capacity, token-based AR models still rely on sequential next-token prediction, leaving inference inherently slow.

\subsection{Efficient Autoregressive Image Generation}
The sequential nature of AR decoding in large-scale image generation models often leads to both memory bandwidth bottlenecks and high inference latency. Existing efforts mainly improve efficiency from two complementary perspectives: compressing the key-value (KV) cache to reduce memory overhead, and reducing sequential decoding steps through parallel decoding.

As AR generation is predominantly memory-bound, a significant line of work focuses on compressing the KV cache to alleviate I/O constraints. Early efforts in language modeling like H2O~\citep{zhang2023h2o} and KIVI~\citep{liu2024kivi} propose efficient eviction and quantization strategies. Subsequent works, such as SnapKV~\citep{li2024snapkv}, identify and retain only the most salient KV pairs. SSD~\citep{jian2025ssd} demonstrates the advantage of decoupling the compression process across different attention heads, thereby achieving superior compression rates.

Another direction reduces the sequential dependency of AR models by enabling multi-token prediction. LPD~\citep{zhang2025locality} improves parallel autoregressive decoding through a locality-aware generation schedule that reduces intra-group dependencies. PAR~\citep{wang2025parallelizedpar} likewise exploits spatial dependency structure by generating weakly dependent distant tokens in parallel while preserving sequential generation for strongly dependent local tokens. ARPG~\citep{li2025autoregressivearpg} enables randomized parallel generation through guided random-order prediction rather than fixed raster-order decoding. ZipAR~\citep{he2024zipar} further adopts a training-free design, exploiting spatial locality to decode multiple image tokens in parallel. Other approaches adapt Speculative Decoding~\citep{leviathan2023fast, chen2023accelerating, cai2024medusa, li2024eagle} and Jacobi Decoding~\citep{santilli2023accelerating, song2021accelerating, kou2024cllms} to autoregressive image generation. Speculative Decoding uses a lightweight draft model to propose multiple candidate tokens for parallel verification by the target model, and LANTERN~\citep{jang2025lantern} improves this paradigm by relaxing the acceptance criterion. In contrast, Jacobi Decoding enables parallel multi-token generation through iterative refinement toward convergence. SJD~\citep{tengaccelerating} extends this idea to sampling-based visual autoregressive generation with a probabilistic acceptance criterion, while GSD~\citep{so2025grouped} further improves efficiency by jointly verifying groups of valid image tokens.

Unlike prior approaches that either require retraining or overlook the spatial structure of images, our method accelerates decoding in a training-free manner by explicitly leveraging spatial locality in image generation. This yields substantial speedup while preserving high visual fidelity.

\section{Method}
\subsection{Preliminary}

\begin{figure*}[t]
    \centering
    \includegraphics[width=\linewidth]{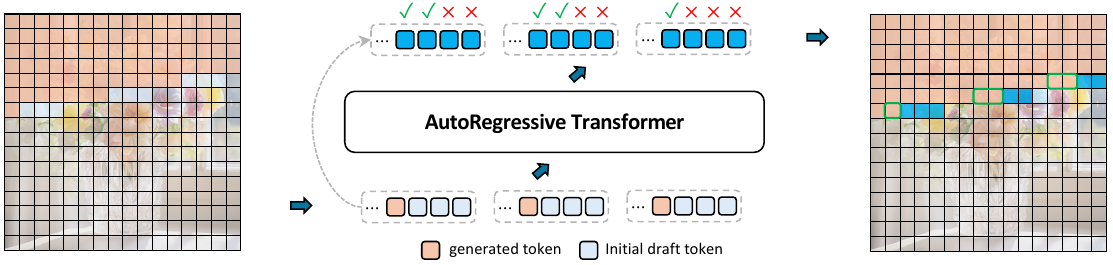}
    \caption{An illustration of one PJD iteration. (Left) Three rows become simultaneously active, each initializing three draft tokens. (Middle) All active rows are processed in one forward pass of the autoregressive transformer, followed by row-wise validation: accepted tokens are committed, while rejected ones are reused as the initial drafts for the next iteration. (Right) Each row’s sliding window advances after validation.}
    \label{fig:iteration}
\end{figure*}

\paragraph{Autoregressive Image Generation.}
The autoregressive image generation pipeline consists of three components: 
a tokenizer that converts continuous images into discrete visual tokens, 
an autoregressive model that generates tokens sequentially, 
and a decoder that reconstructs the image.
Given an image $X \in \mathbb{R}^{H \times W \times C}$, the tokenizer first encodes it into a latent feature map
$Z = f_\psi(X) \in \mathbb{R}^{h \times w \times d}$ (e.g., VQ-VAE~\citep{oord2017vqvae}, VQ-GAN~\citep{esser2021taming}).
Each latent vector $Z_{ij} \in \mathbb{R}^d$ is then replaced by its nearest entry in a learned codebook $\{\mathbf{c}_1,\dots,\mathbf{c}_Q\}$, yielding a quantized embedding map $X_e = (\mathbf{c}_{q_{ij}}) \in \mathbb{R}^{h \times w \times d}$.
Flattening $X_e$ in raster-scan order produces a token sequence $\mathbf{x}=(x_1,\dots,x_L)$ with $L=h\times w$, where each $x_i$ denotes a discrete codebook token associated with one spatial location.
The autoregressive model parameterized by $\theta$ then models this sequence distribution as:
\begin{equation}
    p_\theta(\mathbf{x})=\prod_{i=1}^{L} p_\theta\!\left(x_i \mid x_{1:i-1}\right).
\end{equation}
During inference, it generates one token at a time by sampling ${x}_i \sim p_\theta\!\left(x_i \mid {x}_{1:i-1}\right)$.
These generated tokens are finally mapped back to their corresponding codebook embeddings, reshaped into a spatial grid $\hat{X}_e \in \mathbb{R}^{h \times w \times d}$, and decoded by $g_\phi$ back into pixel space to reconstruct the image $\hat{X} = g_\phi(\hat{X}_e) \in \mathbb{R}^{H \times W \times C}$.

\paragraph{Jacobi Decoding.}
Jacobi decoding reformulates autoregressive generation as a parallel fixed-point iteration~\citep{song2021accelerating}. 
Given a confirmed prefix $\mathbf{x}_{1:t}$ and a future window of size $W$, it initializes a draft sequence $\mathbf{y}^{(0)}=(y^{(0)}_{t+1},\dots,y^{(0)}_{t+W})$ for the ungenerated positions. At iteration $k$, all draft tokens are updated in parallel according to:
\begin{equation}
    \mathbf{y}^{(k+1)}=\mathcal{F}_\theta(\mathbf{y}^{(k)};\mathbf{x}_{1:t}).
\end{equation}
where $\mathcal{F}_\theta$ predicts all positions in parallel using the entire previous draft and the confirmed prefix as context. The iteration proceeds for several rounds until a stopping criterion is met,
such as convergence ($\mathbf{y}^{(k+1)}=\mathbf{y}^{(k)}$~\citep{santilli2023accelerating}) or a confidence-based criteria~\citep{tengaccelerating}. It then extends the confirmed prefix with the longest stable prefix of 
$\mathbf{y}^{(k+1)}$ and shifts the window forward. 
By refining multiple positions in parallel, Jacobi decoding can reduce the number of sequential decoding rounds, potentially improving inference latency without additional training.

\subsection{Parallel Jacobi Decoding}
In prior Jacobi-style decoding methods~\cite{tengaccelerating,so2025grouped}, a 1D sequence of draft tokens is constructed before each iteration and fed into the model to predict token distributions.
In contrast, our method incrementally populates the 2D image grid with new draft tokens during generation, which enables parallel Jacobi-style refinement across rows.

As illustrated in Figure~\ref{fig:iteration} and summarized in Algorithm~\ref{alg:pjd}, each decoding iteration in PJD proceeds through three main stages. 
First, \textbf{Dynamic Token Preparation} identifies eligible rows and constructs new draft tokens based on the available spatial context.
Next, \textbf{Parallel Autoregressive Prediction} performs a single forward pass with a row-causal attention mask to compute the conditional probabilities of all draft tokens simultaneously. 
Finally, \textbf{Probabilistic Token Validation} evaluates whether draft tokens have converged across iterations, selectively accepting or resampling tokens. 

Through this spatially adaptive iteration scheme, PJD achieves efficient parallel refinement while preserving generation quality.

\begin{algorithm}[t]
\caption{Parallel Jacobi Decoding}
\label{alg:pjd}
\begin{algorithmic}[1]
\Statex \hspace{-\algorithmicindent}\textbf{Input:} AR model $p_\theta$, row length $C$, number of rows $R$, draft window size $W$, context threshold $T_{\mathrm{ctx}}$
\State Initialize $k \gets 0,\; p_{ij}^{0} \leftarrow \text{Random}(),\; O \leftarrow \{0\}$

\While{$O \neq \emptyset$}
\State $O_{\text{new}} \leftarrow \emptyset$
\State \textbf{for} $i \in O$:
        \State \hspace{\algorithmicindent} \textbf{if} $i-1 \in O$: $L_i \gets \min\!\big(W,\; |X_{i-1}^{(k)}|-|X_i^{(k)}|\big)$
        \State \hspace{\algorithmicindent} \textbf{else}: $L_i \gets \min\!\big(W,\; C-|X_i^{(k)}|\big)$
        \State \hspace{\algorithmicindent} \textbf{if} {$|X_i^{(k)}| \ge T_{\mathrm{ctx}}$ \textbf{and} $i+1<R$ \textbf{and} $i+1 \notin O$}:
        \State \hspace{\algorithmicindent} \hspace{\algorithmicindent} Add $i+1$ to $O_{\text{new}}$
        \State \hspace{\algorithmicindent} \hspace{\algorithmicindent} $L_{i+1} \leftarrow \min(W, |X_i^{(k)}| - |X_{i+1}^{(k)}|)$
\State$O \leftarrow O \cup O_{\text{new}}$

\State \textbf{parallel for} all $i\in O$ and $j\in[|X_i^k|,|X_i^k|+L_i-1]$:
\State \hspace{\algorithmicindent} $X_{ij}^{k} \sim p_{ij}^{k}(x)$

\State \textbf{parallel for} all $i\in O$ and $j\in[|X_i^k|,|X_i^k|+L_i-1]$:
\State \hspace{\algorithmicindent} $p_{ij}^{k+1} \leftarrow p_\theta(\cdot \mid X^k)$

\State \textbf{for} {$i \in O$}:
    \State \hspace{\algorithmicindent} \textbf{for} {$j = |X_i^{k}|$ \textbf{to} $|X_i^{k}| + L_i - 1$}:
        \State \hspace{\algorithmicindent} \hspace{\algorithmicindent} Sample $u \sim U[0,1]$
        \State \hspace{\algorithmicindent} \hspace{\algorithmicindent} \textbf{if} {$u \le \min\!\left(1, \frac{p_{ij}^{k+1}(X_{ij}^{k})}{p_{ij}^{k}(X_{ij}^{k})}\right)$}: $X_{ij}^{k+1} \leftarrow X_{ij}^{k}$
        \State \hspace{\algorithmicindent} \hspace{\algorithmicindent} \textbf{else}:
        \State \hspace{\algorithmicindent} \hspace{\algorithmicindent} \hspace{\algorithmicindent} \textbf{if} {$i = \min(O)$}: $X_{ij}^{k+1} \sim [p_{ij}^{k+1} - p_{ij}^{k}]_+$
        \State \hspace{\algorithmicindent} \hspace{\algorithmicindent} \hspace{\algorithmicindent} \textbf{break}

\State $k \gets k+1$, \quad $O \gets \{\,i \in O \mid |X_i^{(k)}| < C\,\}$
\EndWhile
\Statex \hspace{-\algorithmicindent}\textbf{Output:} generated image token grid $X$
\end{algorithmic}
\end{algorithm}

\paragraph{Stage 1: Dynamic Token Preparation.}
Leveraging the spatial locality of images, new rows are activated for parallel Jacobi decoding once sufficient upper-row context becomes available.
Inspired by prior work~\citep{he2024zipar}, we measure this context by the number of generated tokens in the preceding row, which we refer to as the \emph{Context Token Count}.
At the beginning of iteration $k$, we first identify the rows that can be activated for Jacobi decoding. For each row $i$, let $c_i^{(k-1)}$ denote the number of generated tokens in row $i$ after iteration $k-1$.
A subsequent row $i+1$ becomes eligible for activation once the preceding row accumulates at least a predefined context threshold $T_{\mathrm{ctx}}$:
\begin{equation}
    O_k = \{\, i+1 \mid c_i^{(k-1)} \ge T_{\mathrm{ctx}},\; i+1 \notin O_{<k} \,\}.
\end{equation}
where $O_{<k}$ is the set of rows activated before iteration $k$.

For each newly activated row $i+1 \in O_k$, we append a sequence of draft tokens of length:
\begin{equation}
    L_{i+1}^{(k)} = 
    \min\!\big(W,\, c_i^{(k-1)} - c_{i+1}^{(k-1)}\big).
\end{equation}
where $W$ is the maximum number of draft tokens appended to a row in one iteration. 
When a row is activated for the first time, we insert an initialization token: for Lumina-mGPT~\cite{liu2024lumina}, we use its end-of-line token, whereas for LlamaGen~\cite{sun2024autoregressive}, we use the token closest to the end of the previous row.
This context-driven activation expands the decoding frontier diagonally across the token grid, triggering new rows only when sufficient context from upper rows is available. This avoids premature activation and maintains spatial coherence during parallel Jacobi decoding.

\begin{figure}[t]
    \centering
    \includegraphics[width=0.75\linewidth]{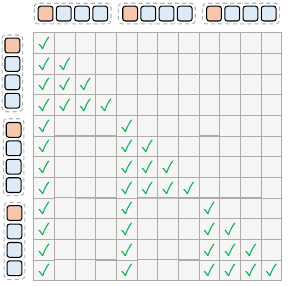}
    \caption{
        Row-causal attention mask in PJD. For every active row, all draft tokens in preceding rows are entirely masked out, allowing visibility only to committed tokens from preceding rows and prior positions within the same row.
    }
    \label{fig:attention_mask}
\end{figure}

\paragraph{Stage 2: Parallel Autoregressive Prediction.}
During prediction, the Transformer computes conditional probabilities for all draft tokens in a single forward pass. This is enabled by a row-causal attention mask (Figure~\ref{fig:attention_mask}), which enforces autoregressive ordering while permitting parallel updates across rows. Tokens in the current row cannot attend to draft tokens in other rows but retain access to committed tokens above. This design preserves valid cross-row context and avoids interference between concurrent draft updates.

\paragraph{Stage 3: Probabilistic Token Validation.}
In this stage, we decide which draft tokens have converged and can be accepted.
In autoregressive image generation, top-k sampling introduces randomness that is essential for maintaining sample diversity. 
Consequently, the deterministic convergence rule used in Jacobi decoding for language models, which accepts a token when it remains identical across iterations, does not apply here.
We instead adopt a probabilistic convergence criterion~\cite{tengaccelerating, leviathan2023fast} that evaluates the stability of each token’s conditional probability across successive iterations.
Specifically, let $x_{ij}$ denote the token at row $i$ and column $j$ of the image token grid. For each token $x_{ij}$ at iteration $k$, we compare the conditional likelihood assigned to it under the current and previous iterations, namely $p_\theta^{(k)}(x_{ij})$ and $p_\theta^{(k-1)}(x_{ij})$.
We draw a random sample $u \sim U[0,1]$ and accept the token if
\begin{equation}
    u \le \min\!\left(1,\,
    \frac{p_\theta^{(k)}(x_{ij})}{p_\theta^{(k-1)}(x_{ij})}
    \right).
    \label{eq:prob_criterion}
\end{equation}
This stochastic rule favors tokens whose likelihood remains stable across iterations, while remaining tolerant to sampling noise.
For the topmost active row, tokens are validated from left to right. 
If the first token fails the criterion in Equation~\ref{eq:prob_criterion}, 
we resample it from a calibrated distribution that emphasizes the probability mass newly increased in the current iteration:
\begin{equation}
x_{ij}^{(k)} \sim
\frac{\max\!\left(0,\, p_\theta^{(k)}(x_{ij}) - p_\theta^{(k-1)}(x_{ij}) \right)}
{\sum_{x^{\prime}} \max\!\left(0,\, p_\theta^{(k)}(x^{\prime}) - p_\theta^{(k-1)}(x^{\prime}) \right)}.
\label{eq:resample}
\end{equation}
After resampling, validation for that row terminates, ensuring that at least one new token 
is finalized in every iteration.

For subsequent rows, we apply the same token-wise test.
Once a token is rejected, that token and all later tokens in the row are deferred to the next iteration.
The rejected token is then updated using the current model prediction to serve as the draft token for the next iteration.
This ensures that rejected tokens carry forward the latest model prediction, avoiding stale drafts while maintaining parallel refinement.

In summary, this probabilistic validation checks \emph{distributional stability} rather than exact token identity, making it naturally compatible with stochastic sampling.
It ensures that each iteration finalizes stable tokens, enabling progressive and reliable refinement across multiple image rows.

\section{Experiments}
\begin{table*}[t]
  \small
  \caption{Quantitative comparison of image generation methods on the MS-COCO dataset. Latency represents the time to generate a single image, and Step indicates the number of steps required. The acceleration factors for both latency and steps are relative to Vanilla AR.}
  \label{tab:quantitative_coco}
  \centering
  {\renewcommand{\arraystretch}{0.88}
  \setlength{\tabcolsep}{4pt}
  \begin{tabular*}{0.9\textwidth}{@{\extracolsep{\fill}}llccccccc@{}}
  \toprule
  \textbf{Model} & \textbf{Configuration} & \textbf{Latency (↓)} & \textbf{Step (↓)} &
  \multicolumn{2}{c}{\textbf{Acceleration (↑)}} & \textbf{FID (↓)} & \textbf{CLIP-Score (↑)} & \textbf{IS (↑)}\\ 
  \cmidrule(lr){5-6}
  & & & & \textbf{Latency} & \textbf{Step} & & \\ 
  \midrule
  \multirow{6}{*}{Lumina-mGPT} 
  & Vanilla AR & 197.16s & 2357 & 1.00× & 1.00× & 30.79 & 31.31 & 32.81\\
  & SJD & 52.97s & 1056 & 3.72× & 2.23× & 30.87 & 31.65 & 32.94\\
  & GSD & 34.36s & 698 & 5.74× & 3.38× & 33.41 & 31.46 & 31.48\\
  & Ours ($c$=16) & 32.91s & 476 & 5.99× & 4.95× & 31.94 & 31.55 & 31.54\\
  & Ours ($c$=11) & 24.78s & 371 & 7.96× & 6.35× & 32.38 & 31.53 & 31.23\\
  & Ours ($c$=9) & \textbf{22.62s} & \textbf{328} & \textbf{8.72×} & \textbf{7.18×} & 32.98 & 31.48 & 31.01\\
  \midrule
  \multirow{5}{*}{LlamaGen-XL} 
  & Vanilla AR & 49.58s & 1024 & 1.00× & 1.00× & 45.02 & 28.59 & 22.11 \\
  & SJD & 27.55s & 631 & 1.80× & 1.62× & 46.18 & 28.58 & 21.83 \\ 
  & GSD & 19.31s & 383 & 2.57× & 2.67× & 47.13 & 28.12 & 20.89 \\ 
  & Ours ($c$=8) & 13.80s & 248 & 3.59× & 4.13× & 46.23 & 28.66 & 21.57 \\
  & Ours ($c$=6) & \textbf{11.84s} & \textbf{213} & \textbf{4.19×} & \textbf{4.81×} & 45.12 & 28.67 & 22.07 \\
  \bottomrule
  \end{tabular*}}
\end{table*}
\begin{table*}[t]
    \small
    \caption{Quantitative comparison of image generation methods on the PartiPrompt dataset. Latency represents the time to generate a single image, and Step indicates the number of steps required. The acceleration factors are relative to Vanilla AR.}
    \label{tab:quantitative_parti}
    \centering
    {\renewcommand{\arraystretch}{0.88}
    \setlength{\tabcolsep}{4pt}
    \begin{tabular*}{0.9\textwidth}{@{\extracolsep{\fill}}llcccccc@{}}
    \toprule
    \textbf{Model} & \textbf{Configuration} & \textbf{Latency (↓)} & \textbf{Step (↓)} &
    \multicolumn{2}{c}{\textbf{Acceleration (↑)}} & \textbf{CLIP-Score (↑)} & \textbf{IS (↑)}\\ 
    \cmidrule(lr){5-6}
    & & & & \textbf{Latency} & \textbf{Step} & \\ 
    \midrule
    \multirow{6}{*}{Lumina-mGPT} 
    & Vanilla AR & 181.43s & 2357 & 1.00× & 1.00× & 31.42 & 21.24\\
    & SJD & 50.61s & 1043 & 3.58× & 2.26× & 32.40 & 21.22\\
    & GSD & 31.36s & 679 & 5.79× & 3.47× & 32.15 & 21.15\\
    & Ours ($c$=16) & 26.05s & 366 & 6.96× & 6.44× & 32.22 & 21.25\\
    & Ours ($c$=11) & 25.46s & 365 & 7.13× & 6.46× & 32.17 & 21.36\\
    & Ours ($c$=9) & \textbf{25.10s} & \textbf{324} & \textbf{7.23×} & \textbf{7.27×} & 32.11 & 21.09\\
    \midrule
    \multirow{5}{*}{LlamaGen-XL} 
    & Vanilla AR & 53.57s & 1024 & 1.00× & 1.00× & 28.86 & 13.91\\
    & SJD & 25.69s & 558 & 2.08× & 1.83× & 28.53 & 13.82\\
    & GSD & 18.06s & 368 & 2.96× & 2.78× & 27.98 & 12.69\\
    & Ours ($c$=8) & 12.89s & 246 & 4.15× & 4.16× & 28.34 & 13.80\\
    & Ours ($c$=6) & \textbf{11.62s} & \textbf{212} & \textbf{4.61×} & \textbf{4.83×} & 28.41 & 12.86 \\
    \bottomrule
    \end{tabular*}}
\end{table*}

\subsection{Evaluation Setups}
\paragraph{Implementation Details.}
We conduct experiments on two representative autoregressive text-to-image models: Lumina-mGPT~\citep{liu2024lumina} and LlamaGen~\citep{sun2024autoregressive}.
For Lumina-mGPT, we use the 7B model to generate $768 \times 768$ images with top-k sampling (k = 2000), temperature $\tau{=}1.0$, and a classifier-free guidance (CFG) scale of 3.0. For LlamaGen, we employ the 7B LlamaGen-XL-Stage-2 model to generate $512 \times 512$ images using top-k sampling k = 1000, temperature $\tau{=}1.0$, and a CFG scale of 3.5.
All experiments are implemented in PyTorch 2.5.1~\citep{paszke2019pytorch} and Python 3.10, and conducted on 8 NVIDIA RTX 4090 GPUs.

\paragraph{Datasets and Metrics.}
We benchmark our method on two widely used text-to-image benchmarks, PartiPrompt~\citep{yu2022scaling} and MS-COCO~\citep{lin2014microsoft}. PartiPrompt includes 1,632 diverse text descriptions covering a broad range of visual concepts, 
while MS-COCO provides natural image-caption pairs. To evaluate open-domain generalization, we randomly sample 5,000 captions from the MS-COCO 2017 validation set.
For image quality evaluation, we use three standard metrics: CLIP-Score~\citep{radford2021learning}, Fréchet Inception Distance (FID)~\citep{heusel2017gans}, and Inception Score (IS)~\citep{salimans2016improved}. 
Generation efficiency is measured by the average number of model inference steps and the wall-clock latency per image, providing a direct quantification of the acceleration achieved by our method.

\paragraph{Comparison Methods.}
We compare our method with three methods: 
(1) Vanilla AR, the standard next-token autoregressive decoding process without acceleration;  
(2) Speculative Jacobi Decoding (SJD)~\citep{tengaccelerating}, a probabilistic parallel decoding algorithm that introduces a sampling-compatible convergence criterion to accelerate AR image generation; and 
(3) Grouped Speculative Decoding (GSD)~\citep{so2025grouped}, a training-free decoding scheme that accelerates generation by evaluating clusters of candidate image tokens rather than the single most likely token.
All methods are evaluated under the same model architecture and sampling settings to ensure a fair comparison.

\subsection{Main Results}
\paragraph{Quantitative Results.}
Table \ref{tab:quantitative_coco} presents quantitative results on the MS-COCO dataset. On Lumina-mGPT, our method reduces decoding steps while maintaining comparable FID, CLIP-Score, and IS. With larger context (e.g., $c{=}16$), it matches GSD in quality with fewer steps, and with smaller context it achieves substantial speedups over both Vanilla AR and GSD. A similar pattern is observed for LlamaGen-XL. Using a reduced context of $c{=}6$, our method decreases the total number of sampling steps by $4.81\times$ relative to Vanilla AR, while still preserving the perceptual quality of generated images. This demonstrates that our approach generalizes effectively across architectures with different capacities and design principles. Table~\ref{tab:quantitative_parti} further reports performance on the PartiPrompt benchmark, where our method continues to show consistent improvements. Across both models and all datasets, our approach yields substantial reductions in latency and steps while preserving image quality.
\begin{figure*}
  \centering
   \includegraphics[width=0.78\linewidth]{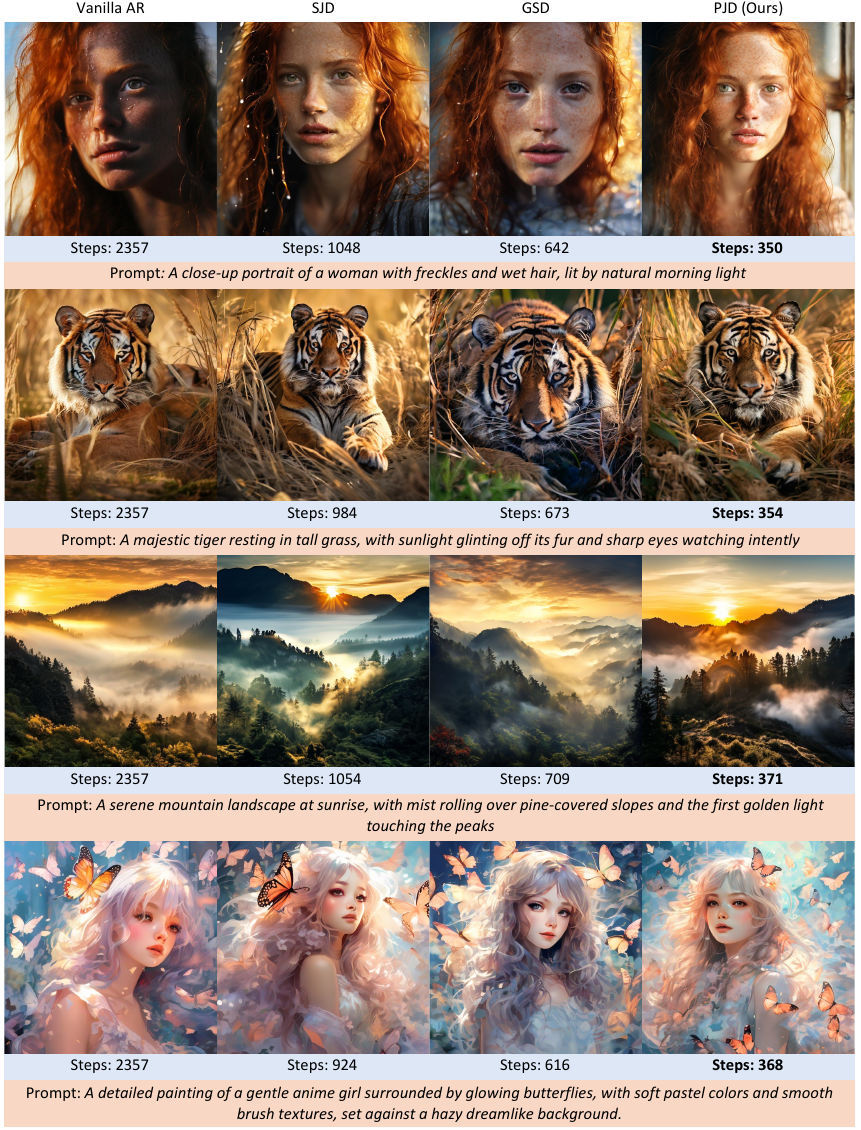}
   \caption{Qualitative comparisons of 768×768 image generation with Lumina-mGPT~\cite{liu2024lumina} across different methods. Our approach accelerates image generation by significantly reducing the number of steps, while maintaining the same level of image quality.}
   \label{fig:qualitative}
\end{figure*}

\begin{figure*}[t]
  \centering
  \begin{minipage}[t]{0.32\linewidth}
    \centering
    \includegraphics[width=\linewidth]{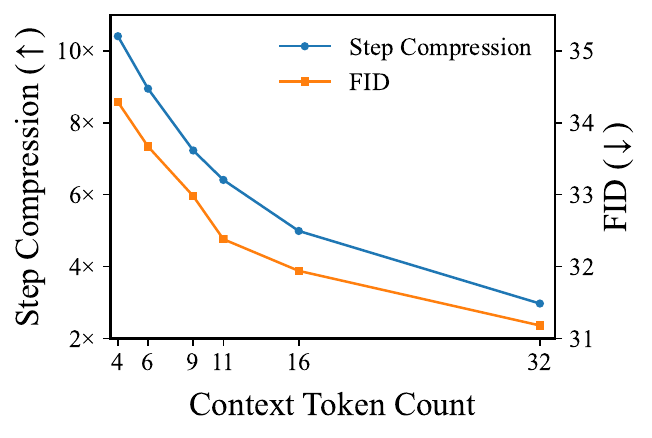}
    \caption{Effect of Context Token Count $c$. A larger $c$ improves image quality (lower FID) but reduces acceleration.}
    \label{fig:context_size}
  \end{minipage}
  \hfill
  \begin{minipage}[t]{0.32\linewidth}
    \centering
    \includegraphics[width=\linewidth]{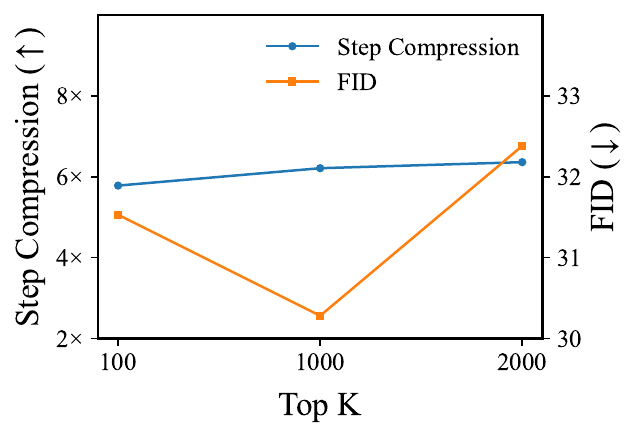}
    \caption{Ablation on top-k sampling. Our method maintains stable acceleration under all tested top-k configurations.}
    \label{fig:topk}
  \end{minipage}
  \hfill
  \begin{minipage}[t]{0.32\linewidth}
    \centering
    \includegraphics[width=\linewidth]{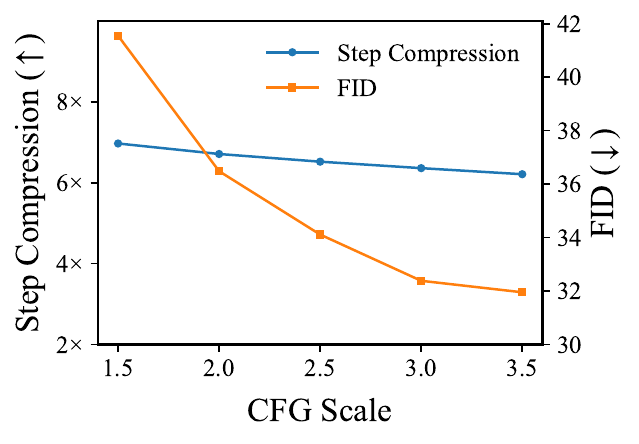}
    \caption{Impact of CFG scale on Step Compression and FID; higher CFG scales yield lower FID while maintaining over $6\times$ acceleration.}
    \label{fig:cfg}
  \end{minipage}

  \label{fig:overall_analysis}

\end{figure*}
\paragraph{Qualitative Results.}
We conduct qualitative experiments under four configurations: Vanilla AR, SJD~\citep{tengaccelerating}, GSD~\citep{so2025grouped}, and our proposed method. 
To comprehensively evaluate image generation performance, we construct four prompt categories covering humans, animals, natural scenes, and cartoon styles, capturing a wide range of visual and semantic diversity. 
All methods adopt identical sampling strategies for a fair comparison. 
The qualitative results, shown in Figure~\ref{fig:qualitative}, demonstrate that our approach consistently produces high-quality and perceptually coherent images across all scenarios. 
Moreover, our method achieves a notable $6.5\times$ inference speedup compared to Vanilla AR.

\subsection{Ablation Study}

\paragraph{Context Token Count.}
We further study the impact of the Context Token Count $c$, which controls when the next row starts decoding during generation. 
We evaluate six settings of $c \in \{4, 6, 9, 11, 16, 32\}$.
As shown in Figure~\ref{fig:context_size}, a larger $c$ provides richer spatial context, leading to improved image quality as reflected by lower FID scores. 
However, the step compression ratio gradually decreases as $c$ increases, revealing a trade-off between speed and fidelity.

\paragraph{Sampling Strategy.}
We evaluate our method under different top-k settings to study the impact of sampling. Since greedy decoding (k = 1) typically yields noticeably degraded images in large-scale text-to-image models, we exclude it from our analysis.

As shown in Figure~\ref{fig:topk}, our method achieves around $6\times$ acceleration across all choices of k, indicating robustness to the sampling budget. The best image quality is achieved at k = 1000, where the FID reaches 30.28, suggesting that a moderate sampling range offers a favorable balance between diversity and fidelity while preserving acceleration.

We also study classifier-free guidance over a range of guidance scales. As shown in Figure~\ref{fig:cfg}, across all CFG scales, our method achieves over $6\times$ step compression, demonstrating stable acceleration. Meanwhile, higher CFG scales progressively reduce FID, indicating improved visual fidelity without compromising efficiency.

\begin{figure}
  \centering
   \includegraphics[width=0.87\linewidth]{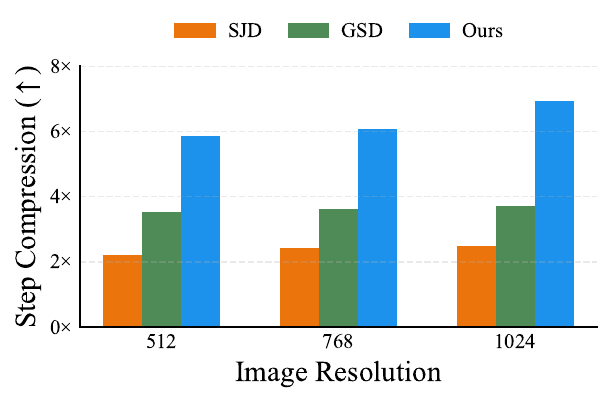}
   \caption{Comparison of step compression ratios across different image resolutions. Our method consistently outperforms SJD and GSD, with the improvement becoming more evident as the image resolution increases.}
   \label{fig:resolution-ablation}
\end{figure}

\paragraph{Image Resolution.}
We evaluate PJD on Lumina-mGPT 7B at three image resolutions to assess its acceleration performance at different generation scales.
For all settings, we fix the Context Coverage Ratio to $0.25$, meaning that decoding for the next row starts once $25\%$ of the tokens in the previous row have been generated.
Empirically, this setting provides a practical trade-off between generation quality and inference efficiency. It corresponds to context token counts of $8$, $12$, and $16$ for the three resolutions, respectively.
As shown in Figure~\ref{fig:resolution-ablation}, PJD accelerates autoregressive generation across all resolutions and outperforms both SJD and GSD by a clear margin. Moreover, its advantage becomes more pronounced at higher resolutions, reaching a step compression ratio of up to \textbf{$6.9\times$} at $1024 \times 1024$. This trend is consistent with the design of PJD, which better exploits 2D spatial parallelism at larger generation scales.

\section{Conclusion}
In this work, we propose \textbf{Parallel Jacobi Decoding (PJD)}, a training-free decoding framework that accelerates autoregressive image generation by enabling spatially parallel refinement within the Jacobi paradigm.
Motivated by the strong spatial locality in autoregressive image attention, PJD schedules updates for positions whose neighborhoods are generated, leading to earlier acceptance.
Extensive experiments across multiple models and datasets demonstrate consistent \textbf{4.8×–6.4×} acceleration without quality degradation, demonstrating that locality-aware scheduling naturally aligns decoding with the spatial structure of images.

\section*{Acknowledgement}
This paper is supported by Young Scientists Fund of the National Natural Science Foundation of China (NSFC) (No. 62506305), Zhejiang Leading Innovative and Entrepreneur Team Introduction Program (No. 2024R01007), Key Research and Development Program of Zhejiang Province (No. 2025C01026), Scientific Research Project of Westlake University (No. WU2025WF003), Chinese Association for Artificial Intelligence (CAAI) \& Ant Group Research Fund - AGI Track (No. 2025CAAI-ANT-13). It is also supported by the research funds of the National Talent Program and Hangzhou Municipal Talent Program.

{
    \small
    \bibliographystyle{ieeenat_fullname}
    \bibliography{main}
}
\clearpage
\setcounter{page}{1}
\maketitlesupplementary
\appendix

\section{Additional Quantitative Results}
\paragraph{Additional Metrics.} We additionally include MLLM-based metrics (VQAScore~\citep{lin2024evaluating} and TIFA~\citep{hu2023tifa}) and human-preference-aligned metrics (ImageReward~\citep{xu2023imagereward} and HPSv2~\citep{wu2023human}). The results in Table~\ref{quantitative_coco} show that, even under these additional quality measurements, our method consistently delivers substantial inference acceleration while maintaining competitive generation quality.
\begin{table*}[t]
  \centering
  \small
  \caption{Quantitative comparison on MS-COCO with additional quality metrics.}
  \label{quantitative_coco}
  \renewcommand{\arraystretch}{0.9}
  \setlength{\tabcolsep}{4pt}
  \begin{tabular*}{0.9\textwidth}{@{\extracolsep{\fill}}lcccccc@{}}
    \toprule
   \textbf{Configuration} &  \textbf{Latency ($\downarrow$)} & \textbf{Step ($\downarrow$)} & \textbf{VQAScore ($\uparrow$)} & \textbf{TIFA ($\uparrow$)} & \textbf{ImageReward ($\uparrow$)} & \textbf{HPSv2 ($\uparrow$)} \\
    \midrule
    Lumina-mGPT   & 197.16s & 2357 & 0.8501 & 0.8545 & 0.7035 & 0.2834 \\
    \textit{w.} SJD & 52.97s & 1056 & 0.8440 & 0.8471 & 0.7004 & 0.2835 \\
    \textit{w.} GSD & 34.36s & 698 & 0.8396 & 0.8469 & 0.6854 & 0.2821 \\
    Ours ($c$=11) & 24.78s & 371 & 0.8399 & 0.8517 & 0.6924 & 0.2825 \\
    Ours ($c$=9) & 22.62s & 328 & 0.8414 & 0.8551 & 0.6782 & 0.2819 \\
    \midrule
    LlamaGen-XL & 49.58s & 1024 & 0.6976 & 0.7317 & -0.2951 & 0.2629 \\
    \textit{w.} SJD & 27.55s & 631 & 0.7192 & 0.7461 & -0.2723 & 0.2636 \\
    \textit{w.} GSD & 19.31s & 383 & 0.6950 & 0.7274 & -0.3230 & 0.2627 \\
    Ours ($c$=8) & 13.80s & 248 & 0.7063 & 0.7458 & -0.2947 & 0.2633 \\
    Ours ($c$=6) & 11.84s & 213 & 0.7059 & 0.7423 & -0.3053 & 0.2633 \\
    \bottomrule
  \end{tabular*}
\end{table*}

\begin{table*}[t]
  \small
  \caption{Quantitative comparison of decoding methods on Janus-Pro across the MS-COCO and PartiPrompt datasets. 
  We report the latency and number of sampling steps required to generate a single 384$\times$384 image, 
  together with acceleration factors relative to Vanilla AR. }
  \label{tab:quantitative_janus}
  \centering
  {\renewcommand{\arraystretch}{0.9}
  \setlength{\tabcolsep}{4pt}
  \begin{tabular*}{0.9\textwidth}{@{\extracolsep{\fill}}llccccccc@{}}
  \toprule
  \textbf{Dataset} & \textbf{Configuration} & \textbf{Latency (↓)} & \textbf{Step (↓)} &
  \multicolumn{2}{c}{\textbf{Acceleration (↑)}} & \textbf{FID (↓)} & \textbf{CLIP-Score (↑)} & \textbf{IS (↑)}\\ 
  \cmidrule(lr){5-6}
  & & & & \textbf{Latency} & \textbf{Step} & & \\ 
  \midrule
  \multirow{6}{*}{MS-COCO} 
  & Vanilla AR & 17.29s & 576 & 1.00× & 1.00× & 33.83 & 32.04 & 32.32\\
  & \textit{w.} SJD & 8.68s & 299 & 1.99× & 1.94× & 34.17 & 32.07 & 32.58\\
  & \textit{w.} GSD & 7.18s & 247 & 2.41× & 2.33× & 34.08 & 32.06 & 32.18\\
  & Ours ($c$=6) & 6.89s & 174 & 2.51× & 3.30× & 33.94 & 32.14 & 32.76\\
  & Ours ($c$=4) & \textbf{5.92s} & \textbf{149} & \textbf{2.92×} & \textbf{3.86×} & 33.96 & 32.15 & 32.48\\ 
  \midrule
  \multirow{6}{*}{PartiPrompt} 
  & Vanilla AR & 17.05s & 576 & 1.00× & 1.00× & - & 32.25 & 19.47\\
  & \textit{w.} SJD & 8.56s & 287 & 1.99× & 2.01× & - & 32.28 & 19.33\\
  & \textit{w.} GSD & 7.02s & 235 & 2.43× & 2.45× & - & 32.30 & 19.43\\
  & Ours ($c$=6) & 6.72s & 162 & 2.54× & 3.56× & - & 32.17 & 19.51\\
  & Ours ($c$=4) & \textbf{5.83s} & \textbf{139} & \textbf{2.92×} & \textbf{4.41×} & - & 32.17 & 19.48\\
  \bottomrule
  \end{tabular*}}
\end{table*}

\begin{table*}[t]
  \small
  \caption{Ablation study on the effectiveness of the Row-Causal Mask (RCM) on the MS-COCO dataset.
  We compare decoding performance on both Lumina-mGPT and LlamaGen-XL, evaluating the impact of
  removing RCM. Incorporating RCM consistently improves sampling efficiency while also
  providing better image quality.}
  \label{tab:ablation_mask}
  \centering
  {\renewcommand{\arraystretch}{0.9}
  \setlength{\tabcolsep}{4pt}
  \begin{tabular*}{0.9\textwidth}{@{\extracolsep{\fill}}llccccccc@{}}
  \toprule
  \textbf{Model} & \textbf{Configuration} & \textbf{Latency (↓)} & \textbf{Step (↓)} &
  \multicolumn{2}{c}{\textbf{Acceleration (↑)}} & \textbf{FID (↓)} & \textbf{CLIP-Score (↑)} & \textbf{IS (↑)} \\
  \cmidrule(lr){5-6}
  & & & & \textbf{Latency} & \textbf{Step} & & & \\
  \midrule
  \multirow{3}{*}{Lumina-mGPT}
  & Vanilla AR & 197.16s & 2357 & 1.00× & 1.00× & 30.79 & 31.31 & 32.81 \\
  & Ours ($c=11$, w/o RCM) & 26.63s & 388 & 7.40× & 6.07× & 32.65 & 31.32 & 31.02 \\
  & Ours ($c=11$) & 24.78s & 371 & 7.96× & 6.35× & 32.38 & 31.53 & 31.23 \\
  \midrule
  \multirow{3}{*}{LlamaGen-XL}
  & Vanilla AR & 49.58s & 1024 & 1.00× & 1.00× & 45.02 & 28.59 & 22.11 \\
  & Ours ($c=6$, w/o RCM) & 12.78s & 227 & 3.88× & 4.41× & 45.67 & 28.50 & 21.26 \\
  & Ours ($c=6$) & 11.84s & 213 & 4.19× & 4.81× & 45.12 & 28.67 & 22.07 \\
  
  \bottomrule
  \end{tabular*}}
\end{table*}

\paragraph{Experiments on Janus-Pro 7B.} We further conduct experiments on Janus-Pro 7B~\citep{chen2025janus} using the MS-COCO~\citep{lin2014microsoft} 
and PartiPrompt~\citep{yu2022scaling} datasets. As shown in Table~\ref{tab:quantitative_janus}, our method achieves the best acceleration performance among all decoding strategies while maintaining competitive image quality on both datasets..

Compared with the results on Lumina-mGPT, the acceleration gains on Janus-Pro 7B are relatively smaller. This performance gap is mainly due to the lower output resolution (384$\times$384) used by Janus-Pro,  which reduces the number of spatial rows available for parallel decoding, thereby limiting the achievable acceleration.
Nevertheless, under this more constrained setting, our method still achieves the best speed–quality trade-off across all methods. Notably, under the setting $c=4$, our approach achieves nearly a 4.41$\times$ step reduction, while maintaining CLIP and IS scores comparable to those of the Vanilla AR baseline.

\section{Additional Qualitative Results}
\begin{figure*}[t]
  \centering
   \includegraphics[width=0.9\linewidth]{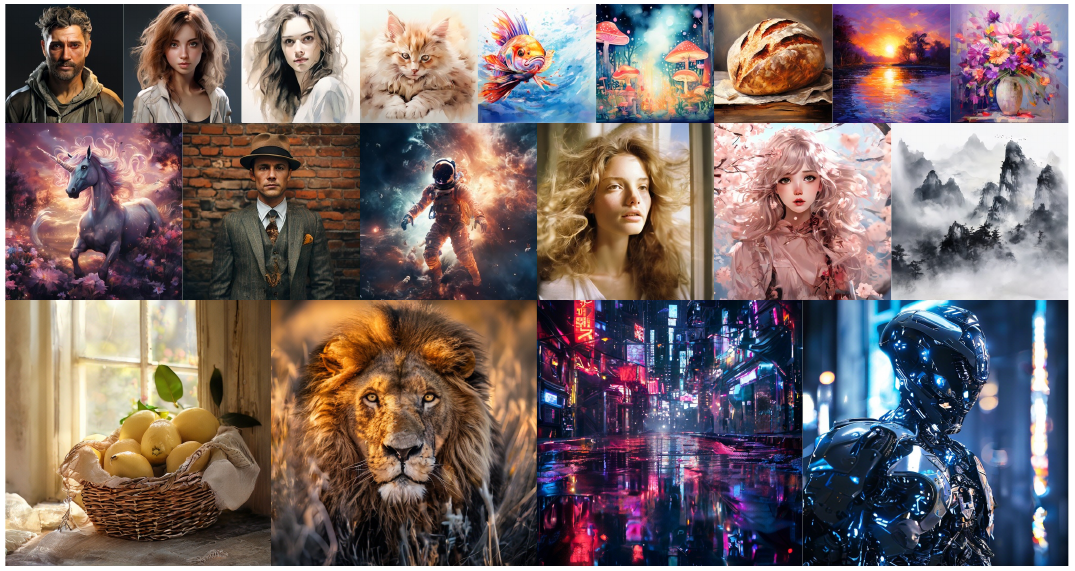}
   \caption{Qualitative results generated by our method on Lumina-mGPT~\citep{liu2024lumina} at different image resolutions. From top to bottom, the rows show images generated at 512$\times$512, 768$\times$768, and 1024$\times$1024 resolutions, respectively. Our method produces high-quality and diverse images across all resolutions, demonstrating strong generative capability and scalability.}
   \label{fig:lumina_qualitative}
\end{figure*}

\paragraph{Qualitative Results on Lumina-mGPT at Multiple Resolutions.}
In Figure~\ref{fig:lumina_qualitative}, we present qualitative results generated by our method on Lumina-mGPT across different output resolutions, including 512$\times$512, 768$\times$768, and 1024$\times$1024. Across all resolutions, our approach consistently produces visually compelling images with rich textures, coherent structures, and diverse artistic styles. Notably, even at higher resolutions where the generative process becomes more challenging due to the increased number of spatial tokens, our method maintains strong fidelity and detail preservation. 

\paragraph{Comparison of Decoding Strategies on LlamaGen.}
We conduct qualitative comparisons using LlamaGen-XL~\citep{sun2024autoregressive} with four decoding strategies: 
Vanilla AR, SJD~\citep{tengaccelerating}, GSD~\citep{so2025grouped}, and our proposed PJD method. 
All methods use the same sampling configuration, including top-k = 1000 and a CFG scale of 3.5, to ensure a fair evaluation across different prompts. For our method, we adopt the setting $c=6$, which balances quality strength and sampling efficiency under this evaluation setup.

\begin{figure*}[t]
  \centering
   \includegraphics[width=0.7\linewidth]{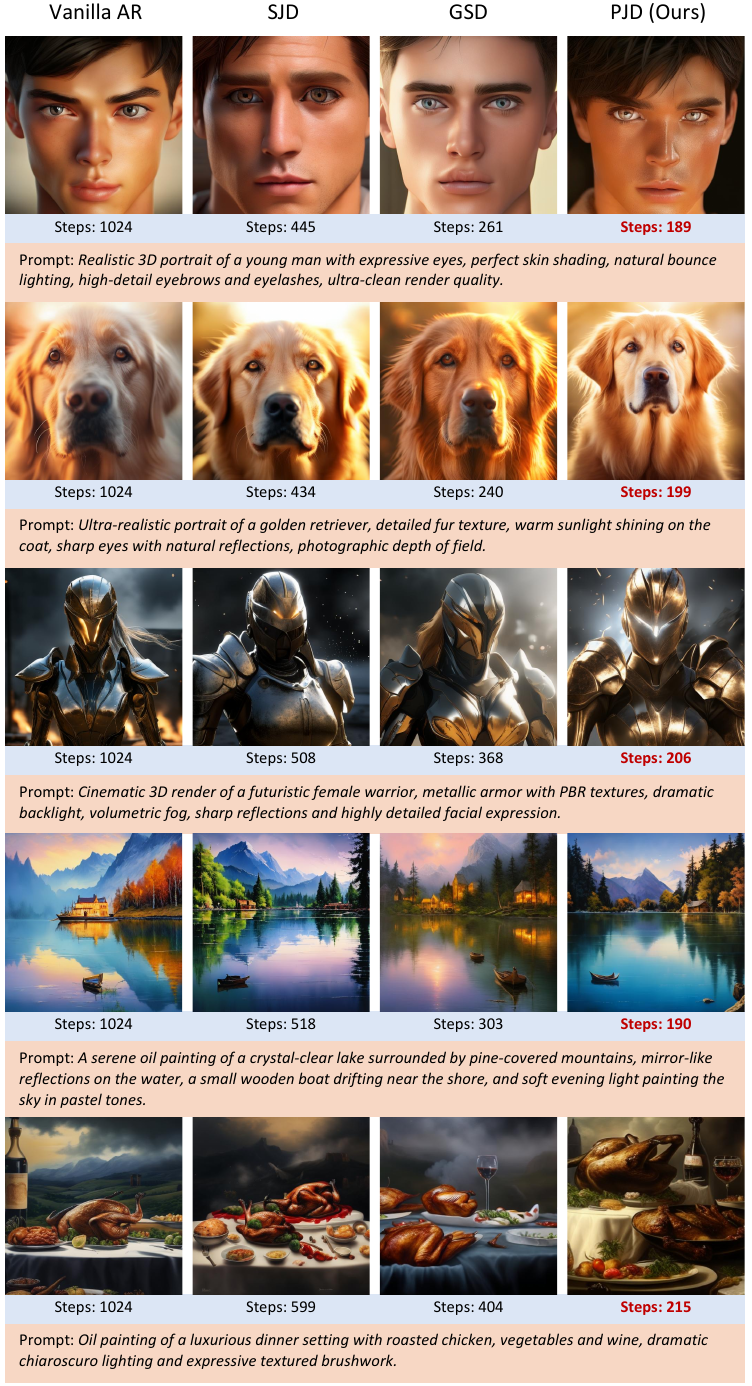}
   \caption{Qualitative comparison of 512$\times$512 image generation results on LlamaGen-XL~\citep{sun2024autoregressive} using four decoding strategies: Vanilla AR, SJD, GSD, and our PJD method. Across all prompts, our approach achieves the fastest generation with the fewest sampling steps.}
   \label{fig:llama_qualitative}
\end{figure*}

As shown in Figure~\ref{fig:llama_qualitative}, our method achieves the 
lowest sampling cost across all prompts, reducing the number of decoding steps by up to 
\textbf{5.4$\times$} compared to the Vanilla AR. 
This reduction directly translates to significantly faster inference while requiring no modification to the underlying LlamaGen model.  Despite the substantial reduction in sampling steps, the image quality of the generated images remains 
comparable to that of the other methods. 

\begin{figure*}[t]
  \centering
   \includegraphics[width=0.9\linewidth]{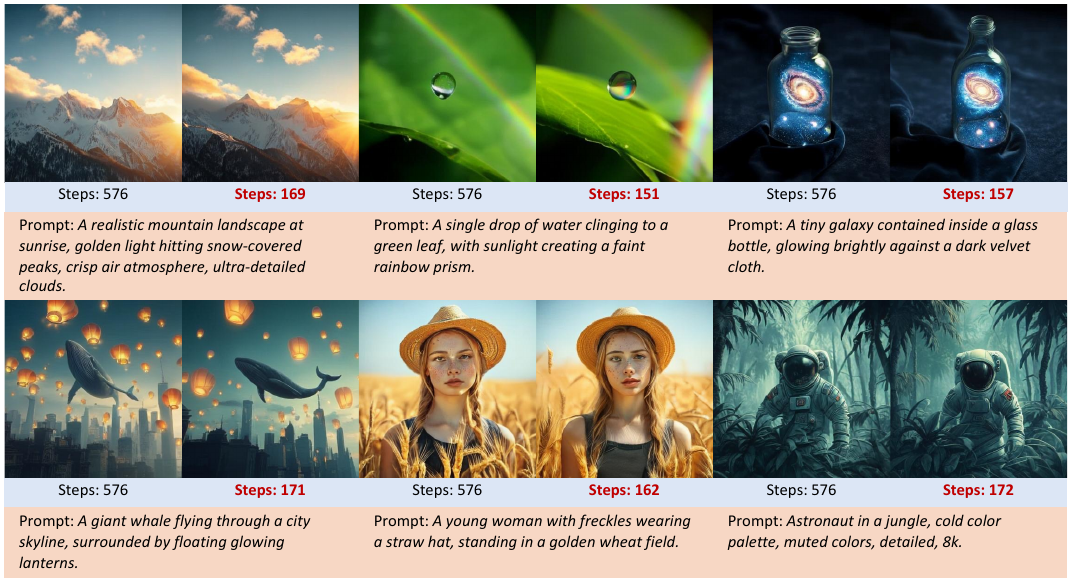}
   \caption{Qualitative comparisons of 384$\times$384 image generation on Janus-Pro~\citep{chen2025janus} across multiple prompts. 
   For each pair, the left image is generated by Vanilla AR and the right image is generated by our method. 
   Our approach significantly reduces the number of sampling steps while preserving comparable image quality.}
   \label{fig:janus_qualitative}
\end{figure*}

\paragraph{Comparison Between PJD and Vanilla AR on Janus-Pro.}
Figure~\ref{fig:janus_qualitative} presents qualitative comparisons of 384$\times$384 image generation on Janus-Pro across multiple prompts. Each pair shows the output of Vanilla AR (left) and our method (right). All methods use the same sampling configuration, including top-k = 1000 and a CFG scale of~5.0. For our method, we adopt the setting $c=4$. Across all examples, our method produces images that are visually comparable to those generated by the full-step Vanilla AR model, despite requiring substantially fewer decoding steps.

\begin{figure*}[t]
  \centering
  \includegraphics[width=0.9\linewidth]{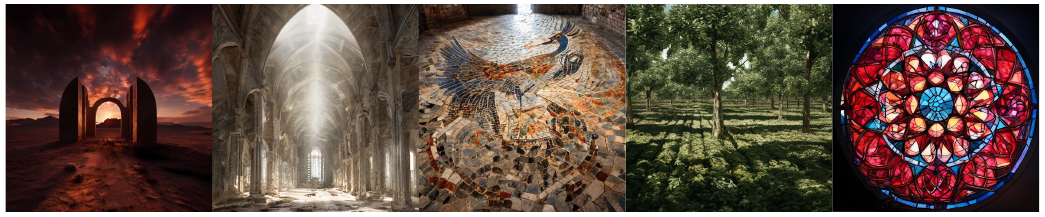}
  \caption{Qualitative weak-locality image generation results of our method on Lumina-mGPT.}
  \label{fig:weak_locality}
\end{figure*}

\paragraph{Qualitative Results on Weak-Locality Images.}
We further provide qualitative examples on weak-locality images, as shown in Figure~\ref{fig:weak_locality}. These examples involve visual structures that rely more heavily on long-range dependencies and global consistency, such as symmetry and cross-region correspondence. Overall, our method is able to generate visually plausible weak-locality images with coherent global layouts. While the overall structure is usually maintained, fine-grained details may exhibit mild degradation. For example, subtle asymmetry or slight inconsistencies may appear in highly structured patterns, as in the rightmost example of Figure~\ref{fig:weak_locality}. 

\section{Ablation Study about Attention Mask}
The ablation study evaluates the effectiveness of the proposed Row-Causal Mask (RCM). 
We compare two settings: using RCM and removing it during decoding. In the w/o RCM setting, each token in a row is allowed to attend to all previously generated tokens as well as all draft tokens produced in the current step. Experiments are conducted on both Lumina-mGPT and LlamaGen-XL using the MS-COCO dataset to assess the impact of RCM on decoding efficiency and image quality.

As shown in Table~\ref{tab:ablation_mask}, disabling RCM consistently results in slower decoding and lower overall performance across both AR models. 
When RCM is enabled, the number of sampling steps is reduced, leading to clear and consistent improvements in decoding efficiency. 
At the same time, image quality also improves: FID, CLIP score, and IS are consistently better than their w/o RCM counterparts. 
This confirms that enforcing row-wise causal dependencies provides more reliable token predictions and reduces error propagation during decoding.

\begin{figure*}[t]
  \centering
  \includegraphics[width=0.9\linewidth]{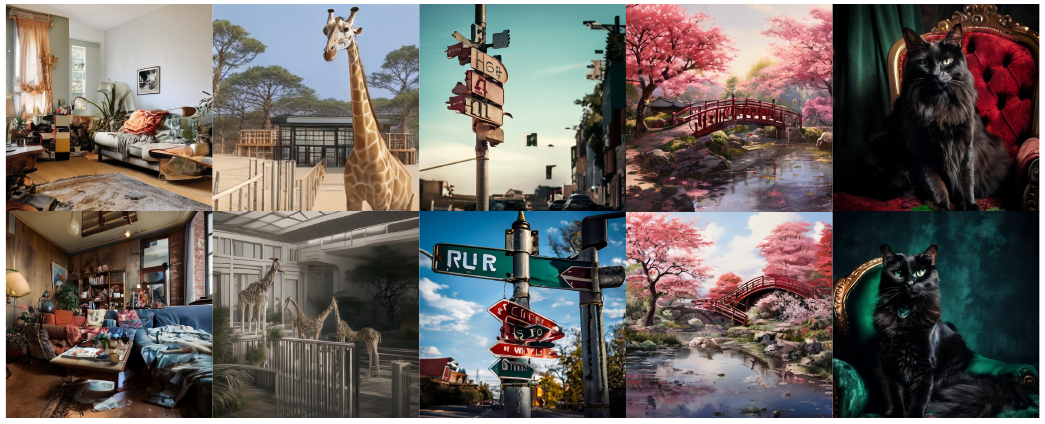}
  \caption{Qualitative comparison with and without the proposed Row-Causal Mask (RCM). The top row shows results with RCM, while the bottom row shows results without RCM.}
  \label{fig:rcm}
\end{figure*}

We also show qualitative comparisons on Lumina-mGPT in Figure~\ref{fig:rcm}, where the top row corresponds to results with RCM and the bottom row corresponds to results without RCM. The results with RCM are generally more natural and visually coherent, while removing RCM tends to produce less reasonable structures and less consistent details.

\section{Analysis on Effectiveness of PJD}
\begin{figure}[H]
  \centering
  \includegraphics[width=1.0\linewidth]{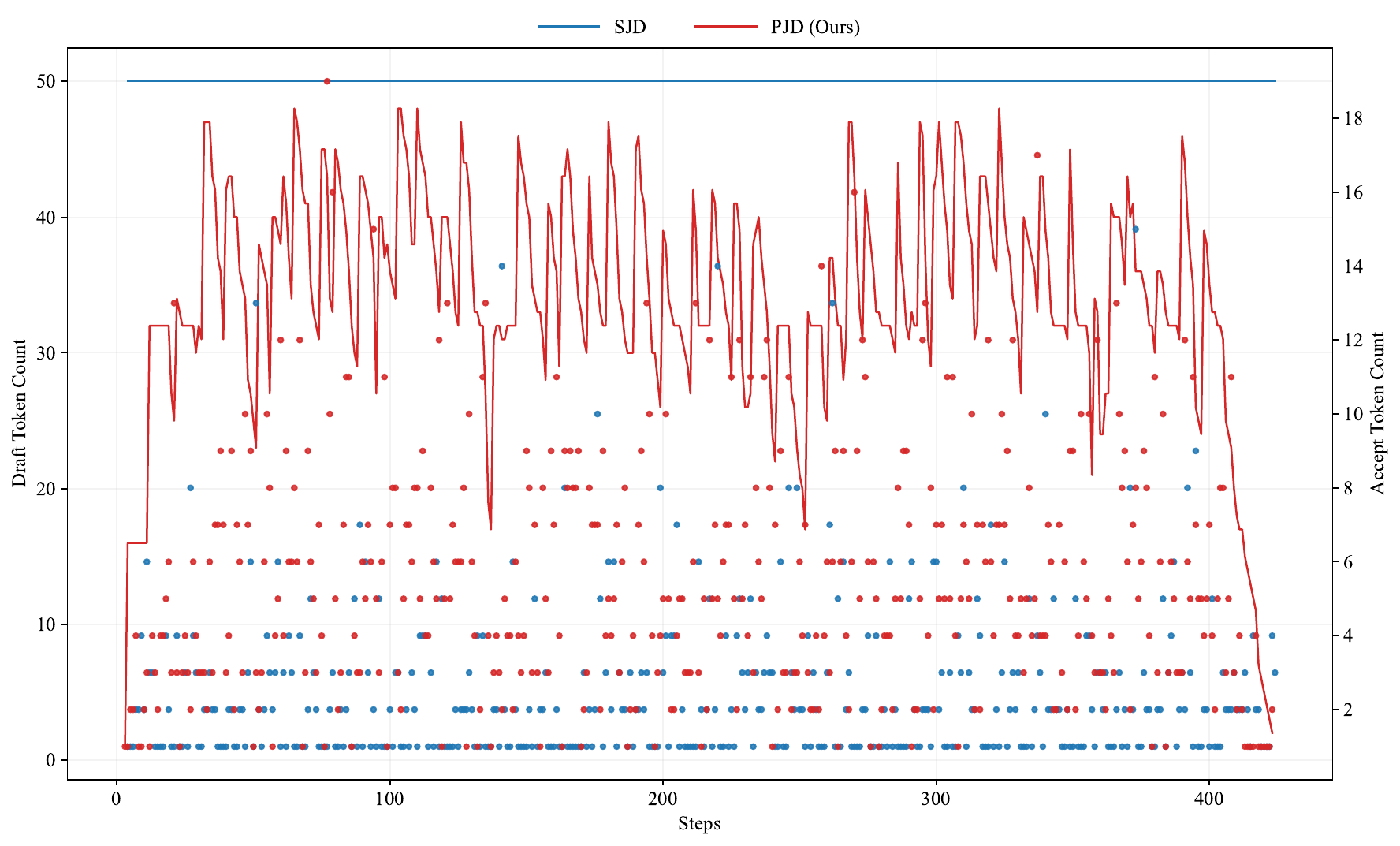}
  \caption{Draft token count (lines) and accepted token count (dots) per decoding step for SJD and PJD.
  PJD uses fewer draft tokens than SJD while achieving a higher accepted token count at each step.} 
  \label{fig:acc}
\end{figure}

Figure~\ref{fig:acc} visualizes the draft and accepted token counts per decoding step for 768$\times$768 image generation. For SJD, the draft token count is fixed to 50 at every step (blue line), whereas our PJD method adaptively adjusts the number of drafted tokens (red line). Importantly, PJD keeps the draft length below 50 across all steps, ensuring that no excessive number of tokens is processed at once. However, the number of accepted tokens per step differs significantly between the two methods. The blue points indicate that SJD accepts only a small fraction of the 50 drafted tokens at each step, 
whereas the red points show that PJD consistently accepts many more tokens per step. This suggests that our adaptive drafting strategy makes much more effective use of the computation, resulting in a higher ratio of accepted tokens to drafted tokens.

\section{Evaluation hardware and VRAM usage} 
All efficiency results are measured on a single RTX 4090 (48GB). We report the peak VRAM usage and the incremental memory cost ($\Delta M$), defined as the maximum additional VRAM during a forward pass. Table~\ref{tab:VRAM} shows that our method accelerates inference with comparable VRAM usage in practice. \\
\begin{table}[t]
  \centering
  \small
  \caption{VRAM usage comparison.}
  \label{tab:VRAM}
  \setlength{\tabcolsep}{6pt}
  \renewcommand{\arraystretch}{0.9}
  \begin{tabular}{lcc}
    \toprule
    \textbf{Configuration} & \textbf{VRAM} & \textbf{$\Delta M$} \\
    \midrule
    Lumina-mGPT   & 16.98G & 0.10G \\
    \textit{w.} SJD  & 17.63G & 0.12G \\
    Ours ($c=11$) & 17.77G & 0.14G \\
    \bottomrule
  \end{tabular}
\end{table}

\section{Prompt Details}
For completeness and reproducibility, we provide the text prompts used to generate 
the four images shown in Figure~\ref{fig:teaser} of the main paper. 
The corresponding prompts, ordered from left to right and from top to bottom, are given as follows:

\begin{enumerate}
    \item \textit{Close-up portrait in the style of Van Gogh, thick brush strokes, vivid colors.}
    \item \textit{Close-up portrait of a red fox with glowing orange fur, richly textured hair strands, warm sunlight reflections, sharp amber eyes, and high-contrast colors creating a vivid, striking appearance.}
    \item \textit{Pixar-quality 3D close-up portrait of a smiling girl, highly detailed hair strands, luminous skin, expressive eyes with detailed reflections, soft rim lighting, premium animation-film rendering, ultra high resolution.}
    \item \textit{Baroque-style oil painting portrait of a young woman with voluminous curly hair, glowing soft light on her face, silky skin texture, rich golden highlights, elegant earrings, luxurious fabric with detailed folds, dramatic dark backdrop, ultra-detailed, masterful brushstroke realism, museum-quality portrait.}
\end{enumerate}

\section{Discussion and Future Work}
While PJD significantly accelerates autoregressive image generation, several limitations suggest promising directions for future work.
First, the current acceptance rule is inherited from speculative decoding and does not explicitly account for 2D spatial structure. Designing convergence rules that incorporate local spatial consistency or region-level stability may further improve distributional faithfulness. Second, although PJD dynamically adjusts the number of draft tokens, extremely large draft regions may challenge GPU parallelism or memory capacity. Exploring hardware-aware scheduling or adaptive token grouping could help ensure scalability at very high resolutions. Overall, PJD highlights the benefits of aligning autoregressive decoding with image structure, and further investigation into acceptance rules, hardware scalability, and generalized 2D generation strategies may enable even faster generation while preserving high image quality. More broadly, recent multimodal reasoning studies~\cite{feng2025can,huang2025vision,feng2025rewardmap} suggest that structured visual understanding and learning-based optimization may also provide useful perspectives for future extensions of efficient visual generation.

\end{document}